\documentclass[11pt]{article}
\usepackage{coling2020}
\usepackage{times}
\usepackage{url}
\usepackage{latexsym}
\usepackage{graphicx}
\usepackage{color}
\usepackage{amsmath,amsfonts,amsthm}
\usepackage{booktabs}
\usepackage{caption}
\colingfinalcopy 

\title{Bi-directional Cognitive Thinking Network for Machine Reading Comprehension}

\author{
	Wei Peng\textsuperscript{1,2}, Yue Hu\textsuperscript{1,2}\thanks{{} {} Corresponding author.}, Luxi Xing\textsuperscript{1,2} \\
	\bf Yuqiang Xie\textsuperscript{1,2}, Jing Yu\textsuperscript{1,2} ,Yajing Sun\textsuperscript{1,2}, Xiangpeng Wei\textsuperscript{1,2}\\
	\textsuperscript{1}Institute of Information Engineering, Chinese Academy of Sciences, China \\
	\textsuperscript{2}School of Cyber Security, University of Chinese Academy of Science, China \\
	{\tt \{pengwei,huyue,xingluxi,xieyuqiang\}@iie.ac.cn} \\
	{\tt \{yujing,sunyajing,weixiangpeng\}@iie.ac.cn}
}
\date{}

\begin{document}
	\maketitle
	\begin{abstract}
		We propose a novel Bi-directional Cognitive Knowledge Framework (BCKF) for reading comprehension from the perspective of complementary learning systems theory. It aims to simulate two ways of thinking in the brain to answer questions, including reverse thinking and inertial thinking. To validate the effectiveness of our framework, we design a corresponding Bi-directional Cognitive Thinking Network (BCTN) to encode the passage and generate a question (answer) given an answer (question) and decouple the bi-directional knowledge. The model has the ability to reverse reasoning questions which can assist inertial thinking to generate more accurate answers. Competitive improvement is observed in DuReader dataset, confirming our hypothesis that bi-directional knowledge helps the QA task. The novel framework shows an interesting perspective on machine reading comprehension and cognitive science.
	\end{abstract}
	
	\section{Introduction}
	\label{intro}
	
	%
	%
	
	Machine Reading Comprehension (MRC) has made significant strides with an array of neural models rapidly approaching human parity on some benchmarks such as SQuAD \cite{DBLP:conf/emnlp/RajpurkarZLL16}. However, existing methods are still in their infancy at the level of cognitive intelligence. Recently, brain science and psychology provide an important basis for the development of brain-like computing and the simulation of human perception, thinking, understanding, and reasoning abilities \cite{sharp2017cognitive}.
	
	Thinking is the generalization and indirect reflection of the human brain on the nature, interrelationships and internal regularities of objective things \cite{wing2006computational}. Two types of thinking are complementary in psychology: \textit{inertial} thinking — from a previous to a subsequent stimulus — and \textit{reverse} thinking — from a subsequent to a previous stimulus \cite{krutetskii1976psychology}. Inertial thinking \cite{DBLP:journals/corr/abs-1803-00158} is a conventional way of thinking, which thinks and solves problems from the previous ideas.
	Reverse thinking \cite{krutetskii1976psychology} is a creative way of thinking, opposite to the inertial thinking. 
	Specifically, in the MRC task, the two types of thinking can be regarded as a process which reasons from questions (answers) to answers (questions). For example, as shown in Fig. \ref{fig:exp0}, we can get the answer easily by locating the entities \textit{pregnant wowen} and \textit{loquat}. Contrarily, the generative question, which can be reasoned by reading the answer and passage, describes two aspects, including \textit{can pregnant women eat loquat} and \textit{what is the benefit to eat loquat for pregnant women}. We hope that this ability of reverse reasoning can improve performance on reading comprehension tasks.

	Previous methods \cite{DBLP:conf/iclr/Wang017a,DBLP:conf/iclr/SeoKFH17,DBLP:journals/corr/TanWYLZ17,DBLP:conf/acl/NishidaSNSOAT19} only consider a obverse logical relationship, which is based on the given question and the passage. They ignore the reverse relationship between the given passage and the answer. Although the work \cite{DBLP:journals/corr/WangYT17} proposes a joint model that both asks and answers questions, it couples all the knowledge rather than decopuling modules, which is consistent with the concept of psychology. Similarly, we hypothesize that the ability of reverse reasoning can help models achieve better QA performance. This is motivated partly by observations made in psychology that devising questions while reading can 
	help students improve in reader-based processing of text \cite{singer1982active}.
	
	
	Therefore, insights into solutions to the problem can be gained from human cognitive processes. Complementary Learning Systems Theory (CLST) \cite{marr1991simple,mcclelland1995there,kumaran2016learning,DBLP:journals/corr/abs-1912-05877} suggests that the human brain contains complementary learning systems that support the simultaneous use of many sources of information as we seek to understand an experienced situation. One of the systems acquires an integrated system of knowledge gradually through interleaved learning, including our knowledge of the meanings of words, the properties of frequently-encountered objects, and the characteristics of familiar situations. It is just like inertial thinking that learns relationships between different things in the real world for a long time. The other system is a fast learning system similar to reverse thinking, which is targeted to focus on stimulating and enhancing infrequently-utilized circuit areas in the brain from another unusual perspective.
	
	In this paper, we propose the \textbf{Bi-directional Cognitive Knowledge Framework (BCKF)}. And the corresponding \textbf{Bi-directional Cognitive Thinking Network (BCTN)} is designed to validate the effectiveness of the reverse thinking, as shown in Fig.~\ref{fig:model}, which will be introduced in detail in Section \ref{sec:model}.
	%
	
	
	\begin{figure}[tp]
		\framebox{
			\parbox{0.96\textwidth}{
				\small
				\textbf{\textcolor{blue}{Example 1} } \newline
				\textbf{Passage:}  Loquat is a kind of southern fruit with a unique flavor and is loved by people. Loquat has a unique flavor and is rich in nutrients. Pregnant women can eat it. It contains protein, fat, fruit acid, fructose and calcium, phosphorus, iron, sodium, potassium and other minerals. Pregnant women eating loquat can increase appetite, relieve heat and relieve thirst. Loquat can stimulate the secretion of digestive glands, and has a good effect on increasing appetite, helping digestion and absorption, and quenching thirst and relieving heat ...
				\newline
				\textbf{Q:} Loquat, can pregnant women eat it ? \newline
				\textbf{A:} {Pregnant women can eat loquat and pregnant women eating loquat can increase appetite, relieve heat and relieve thirst.} \newline
			}
		}
		\caption{An example in DuReader dataset.} 
		\label{fig:exp0}
	\end{figure}

	
	
	The contributions can be summarized as follows:
	
	\begin{itemize}
		\item From the perspective of Complementary Learning Systems Theory, we propose a Bi-directional Cognitive Knowledge Framework (BCKF) and corresponding Bi-directional Cognitive Thinking Network (BCTN), which simulate the connection of neural circuits in the brain, and determine the stimulus intensity of reverse thinking in memory based on the gate mechanism. 
		\item The proposed model, with the Gate-Reverse Thinker and No-Gate-Inertial Thinker, has the ability to reverse reasoning questions which assists inertial thinking to generate more accurate answers in a decoupling fashion.
		\item We conduct extensive experiments and experimental results show that our proposed BCTN has strong competitiveness with other models. The ablation study validates the importance of our different modules.
		
	\end{itemize}
	
	\section{Related Work}
	\label{related}
	
	\subsection{MRC Datasets}
	Recently, with the released datasets, MRC tasks have attracted significant amounts of attention. At first, the dataset was cloze-style \cite{DBLP:conf/emnlp/XieLDH18} which let the machine find an entity to fill the blank, and then was multiple choice \cite{DBLP:conf/emnlp/LaiXLYH17,DBLP:conf/emnlp/RichardsonBR13} and extractive \cite{DBLP:conf/emnlp/RajpurkarZLL16,DBLP:conf/emnlp/ChoiHIYYCLZ18}. After that, datasets made a transition from extractive task to generative task \cite{DBLP:conf/nips/NguyenRSGTMD16,DBLP:conf/acl/HeLLLZXLWWSLWW18}.
	
	\subsection{MRC Models}
	Mainstream studies \cite{DBLP:conf/iclr/Wang017a,DBLP:conf/iclr/SeoKFH17,DBLP:conf/iclr/XiongZS17} treat machine reading comprehension as extracting answer span from the given passage, which is obtained by predicting the probability distribution of the start and end position of the answer. Some models \cite{DBLP:journals/corr/TanWYLZ17,DBLP:conf/acl/NishidaSNSOAT19} focus
	on generating answers from the question and multiple passages with an extraction-then-synthesis framework or a multi-style abstractive summarization approach. Another inspiring work is from \newcite{DBLP:journals/corr/WangYT17}, where the authors propose to use a sequence-to-sequence framework that encodes the document and generates a question (answer) given an answer (question). However, the model, mixing the bi-directional knowledge with the same module, is hybrid and not decoupled, which is inconsistent with cognitive psychology \cite{krutetskii1976psychology}. Our Bi-directional Cognitive Thinking Network does not consider them simultaneously; instead decouples in a bi-directional way of thinking that stemmed from cognitive psychology.
	
	\subsection{Pre-trained Language Model}
	Another interesting study is pre-trained language models which make a great contribution to a wide range of MRC tasks. 
	Previous work \cite{DBLP:conf/acl/RuderH18,DBLP:conf/naacl/PetersNIGCLZ18} adapt traditional language models to improve downstream tasks. Recently, a significant milestone is the BERT \cite{DBLP:conf/naacl/DevlinCLT19}, which gets new state-of-the-art results on eleven natural language processing tasks.
	Then, some work \cite{DBLP:journals/corr/abs-1906-08237,DBLP:journals/corr/abs-1907-11692} such as XLNet and RoBERTa introduce more data and bigger models for better performance.
	However, it is difficult to effectively execute them on resource-restricted devices, \newcite{DBLP:journals/corr/abs-1909-11942} proposes ALBERT 
	to reduce memory consumption and increase training speed. In this paper, we focus on the RoBERTa models to encode our contextual information and regard them as baselines.

	\section{Bi-directional Cognitive Knowlwdge Framework}
	\label{clst}
	Motivated by Complementary Learning Systems Theory (CLST) \cite{marr1991simple,mcclelland1995there,kumaran2016learning,DBLP:journals/corr/abs-1912-05877}, we proposed the Bi-directional Cognitive Knowledge Framework (BCKF). As shown in Fig. \ref{fig:model} (a), the blue box contains the neocortical system which is organized around a set of inputs. The red box is the medial temporal lobes (MTL) system where the blue ovals (Fusion System, Inertial Thinker, Reverse Thinker, Reasoner and Gate Controller) represent the different modules which are directly or indirectly associated with the orange ovals, defined as a neutral pool containing a small amount of information such as visual and linguistic inputs. Green arrows represent learned connections between different blue ovals, which bind the elements of this embedding together for later reactivation. Green dotted lines indicate that the bi-directional thinkers contain the reasoning module. Blue arrows represent information transfer between different systems. The red and blue circular arrows represent self-learning and self-updating. And the controller determines the stimulus intensity of reverse thinking in memory in order to make different decisions in different situations. Finally, the understanding system guides the behavior of the model and its understanding of the language by combining inertial thinking and reverse thinking.
	
	
	\section{Model}
	\label{sec:model}
	
\begin{figure}[t]
		\centering
		\includegraphics[width=0.93\linewidth]{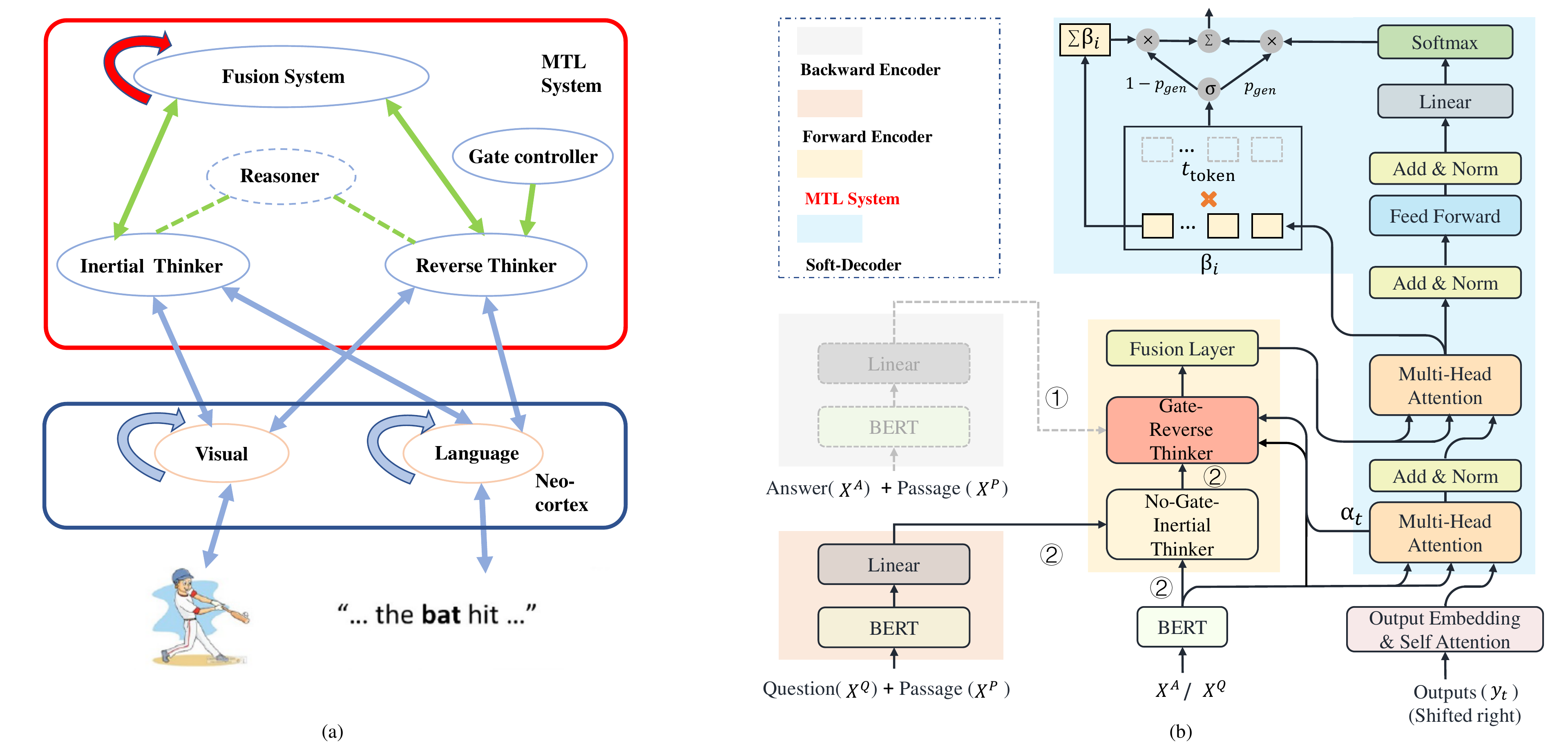}
		\caption{An overview of the proposed method: (a) the framework of our \textbf{Bi-directional Cognitive Knowledge} which consists of neocortical system and medial temporal lobe system (MTL), the arrows indicate the flow of thinking, (b) \textbf{Bi-directional Cognitive Thinking Network}, \textcircled{\small{n}} indicates the different training stages. The gray dotted line indicates that it only exists in the first stage of training. }
		\label{fig:model}
	\end{figure}

	Following the overview in Fig. \ref{fig:model}, the proposed model that derives from our Bi-directional Cognitive Knowledge Framework consists of the following modules, and the training of the model contains two stages.
	
	The Backward Encoder models interactions between the answer and passage during the {first stage} (Backward Encoder $\rightarrow$ Gate-Reverse Thinker $\rightarrow$ Fusion Layer $\rightarrow$ Soft Decoder), called \textbf{Reverse Thinking Training} (\S\ref{Reverse}).
	
	The Forward Encoder, similar to the backward encoder with different parameters and inputs, retrains with the given passage and question during the {second stage} (Forward Encoder $\rightarrow$ No-Gate-Inertial Thinker $\rightarrow$ Gate-Reverse Thinker $\rightarrow$ Fusion Layer $\rightarrow$ Soft Decoder), called \textbf{Retraining with Inertial Thinking} (\S\ref{retain}).
	
	The Medial Temporal Lobes (MTL) System contains Gate-Reverse Thinker, No-Gate-Inertial Thinker and Fusion Layer. The Gate-Reverse Thinker learns the reverse connections of the neurons from the negative side and determines the stimulus intensity of reverse thinking in memory. The No-Gate-Inertial Thinker builds the obverse relationship. And the Fusion Layer combines the bi-directional knowledge to prepare for decoding.
	
	The Soft Decoder outputs an answer (question) sentence with soft-gated pointer-generator copying \cite{DBLP:conf/acl/SeeLM17} to synthesize the distribution on vocabulary and the distribution of tokens from the source input into a single output distribution.
	
	\subsection{Problem Formulation}
	The goal of our task could be formulated as follows. Given a question $X^Q = \{x^q_0, \ldots, x^q_{M-1}\}$ and a passage $X^{P} = \{x^{p}_0, \ldots, x^{p}_{N-1}\}$, the corresponding answer with $K$ words $X^A = \{x^a_0, \ldots, x^a_{K-1}\}$, our model predicts an output sequence $Y = \{y_0, \ldots, y_{T-1} \}$ conditioned on the MTL system that combines the bi-directional knowledge between the passage and question (answer). The training of the model is divided into two stages. In the first stage, the model infers the question based on the passage and answer and it learns the reverse connections of the neurons from the negative side. Then, the model considers reverse thinking and retrains with the given passage and question in a bi-directional schedule of thinking.
	During the prediction, the model combines the bi-directional knowledge to answer the question.
	
	\subsection{Reverse Thinking Training}
	\label{Reverse}
	
	In this section, we train the Gate-Reverse Thinker with the answer and passage where the reserved parameters are regarded as the connection of the reverse circuit in the brain. And the gate is similar to the controller in Fig. (\ref{fig:model}) (a) which determines the stimulus intensity of reverse thinking in memory to make different decisions in different situations. Finally, the decoder infers a question based on the answer.
	
	{\bf Backward Encoder}: Following the implementation of BERT \cite{DBLP:conf/naacl/DevlinCLT19}, we add the special classification embedding ([CLS]) which encodes entailment information between the two sentences and separate the answer $X^A$ and passage $X^P$ with a special token ([SEP]). The total length of the input is $L=$ ($K + N + 3$), where $K$ and $N$ are the length of the answer and passage, respectively. In order to review the answer to locate the answer-relevant semantic information, we encode the answer separately and get a pure answer vector ${V}$ with $(K + 2)$ tokens $V = \{v_0, \ldots, v_{(K+1)}\}$, as:
	
	\begin{align}\label{eq:bert}
	U = {\rm Linear_1}({\rm BERT_1}({\rm [CLS]}, &X^, {\rm [SEP]}, X^P, {\rm [SEP]}))  \\
	\label{V}
	V = {\rm BERT_1}({\rm [CLS]}&, X^A, {\rm [SEP]}) \\
	\label{v}
	u_{cls} = u_0,\; \tilde{V}& = {v}_{0}
	\end{align}
	%
	where $U \in \mathbb{R}^{ L \times h}$ and $V \in \mathbb{R}^{ (K+2) \times h}$, $u_{cls}$ and $\tilde{V}$ denote the representations of the first token of $U$ and $V$. ${\rm Linear_1}$ is a fully connected layer.
	
	\begin{figure}[t]
		\centering
		\includegraphics[width=0.9\linewidth]{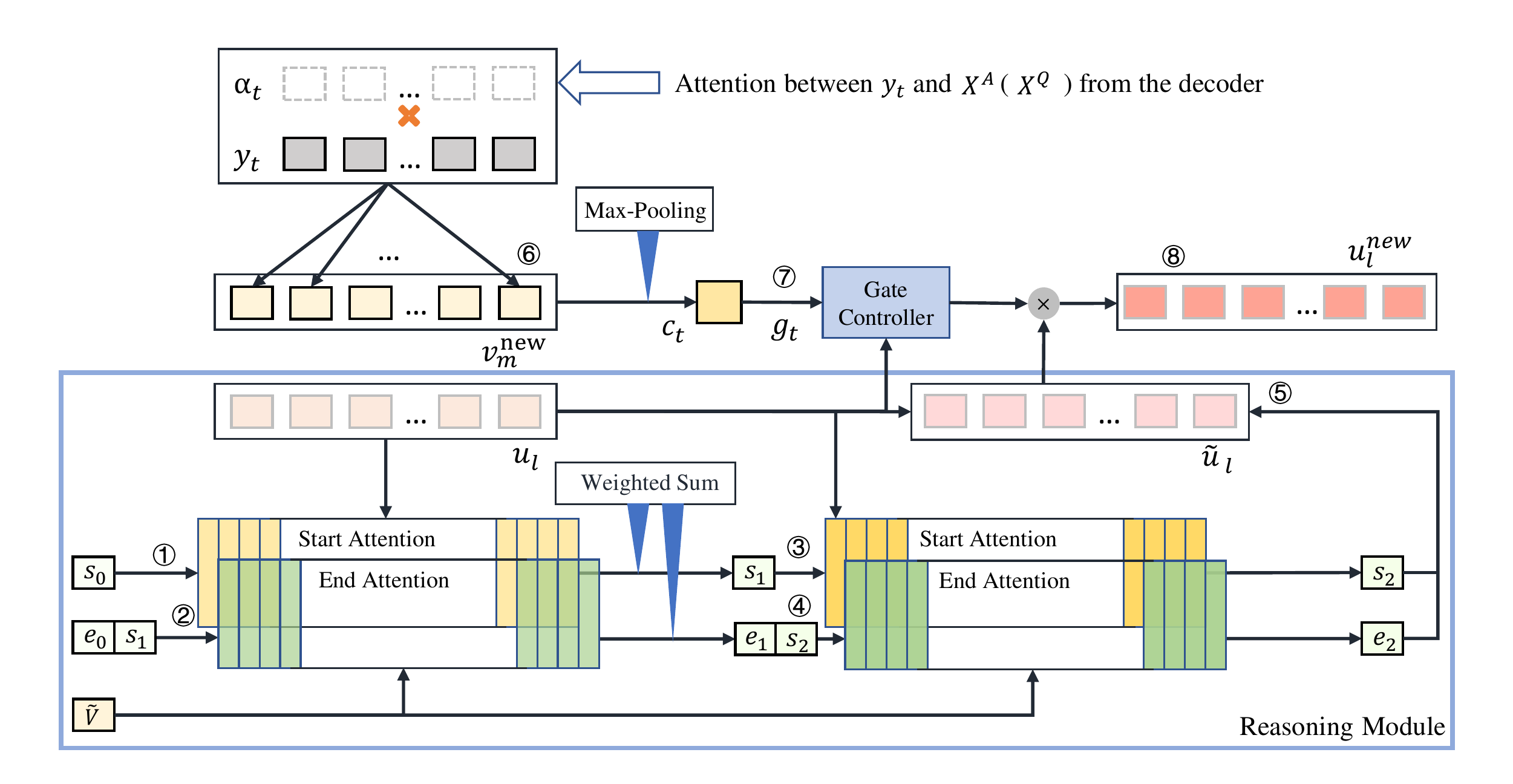}
		\caption{Illustration of the Gate-Reverse Thinker which consists of two parts, Gate Controller and Reasoning Module(blue \fbox{box}). The shade of the color (orange, green and pink) indicates that the status is constantly updated. \textcircled{\small{n}} indicates the order of processes.}
		\label{fig:gate}
	\end{figure}
	
	{\bf Gate-Reverse Thinker}: As Fig. \ref{fig:gate}, the reasoning module (blue \fbox{box}) contains a stack of reasoning blocks which consist of a start (orange) and an end (green) sub-block. These two sub-blocks have a timing dependency, namely, the calculation of the end sub-block needs to consider the result of the start sub-block. The blocks of reasoning module imitate the process of human thinking and dig the relationships constantly between the $U$ and $V$ through multiple steps of reasoning. $ s_j $ and $ e_j $ are the start and end reasoning vectors during the $j$-th reasoning step which can be considered as hidden states to enhance the representation of $U$.
	The final reasoning vectors $ s_2 $ and $ e_2 $ fuse all possible reasoning fragments based on relevance to the answer (or question). Besides, the thinker calculates the gate $g_t$ on the condition of decoded tokens to determine the stimulus intensity of reverse thinking in memory.
	
	Specifically, the module first concatenates $U$ and $s_j$ (Fig \ref{fig:gate} - \textcircled{\small{1}}), we let $s_j$ repeat $L$ times for the consistent dimensions, which followed by \newcite{DBLP:conf/iclr/Wang017a}, as:
	
	\begin{equation}\label{eq:start}
	O^j = {\rm ReLU}(W_{s} [s_j \otimes \mathbf{e}_L; U^\top])
	\end{equation}
	where $W_{s} \in \mathbb{R}^{h \times 2h}$, and $j = \{0,1,2...\}$ means the $j$-th reasoning step, when $j=$ 0, we use a randomly initialized vector $s_0 \in \mathbb{R}^{h}$ and $e_0 \in \mathbb{R}^{h}$ as start and end reasoning vectors, respectively. $(\cdot \otimes \mathbf{e}_L)$ produces a matrix by repeating the vector on the left for $L$ times.
	
	To get the semantic segments that the answer (question) attends, the module rereads it by using the pure answer (question) vector $\tilde V$. Then it computes a start probability distribution $\gamma^j$ over the entire context which will be continuously updated during the reasoning process, as:
	
	\begin{equation}\label{eq:p_start}
	\gamma^j = {\rm Softmax}(\tilde V^\top O^j)
	\end{equation}
	where $\gamma^j \in \mathbb{R}^{L}$, $\tilde V$ and $O^j$ can obtain from Eqn. (\ref{v}) and (\ref{eq:start}).
	
	Then, we use the start probability distribution $\gamma^j$ to obtain the updated start reasoning vector $s_{j+1}$ (Fig. \ref{fig:gate} - \textcircled{\small{3}}) as a guider for latter reasoning. 
	
	\begin{equation}\label{equ:start_new}
	s_{j+1} = \sum^L_{l}\gamma_l^j o_{l}^j
	\end{equation}
	where $o_{l}^j$ and $\gamma_l^j$ are calculated in Eqn. (\ref{eq:start}) and Eqn. (\ref{eq:p_start}), respectively.
	
	The probability distribution $\gamma^j$ and the reasoning vector $s_{j+1}$ of the start position have been calculated. Considering the temporal relationship between the end position and the start position, we introduce $s_{j+1}$ into the calculation of $e_{j+1}$ (Fig. \ref{fig:gate} - \textcircled{\small{4}}). Similarly, the end probability distribution $\eta^j$ and the end reasoning vector $e_{j+1}$ is computed by:
	
	\begin{align}\label{equ:6}
	Z^j = {\rm ReLU}(W_{e} [s_{j+1} &\otimes \mathbf{e}_L; e_j \otimes \mathbf{e}_L; U])	\\
	\eta^j = {\rm Softmax}(&\tilde V^\top Z^j)	\\
	e_{j+1} = \sum^L_{l} &\eta_{l}^j z_{l}^j	
	\end{align}
	where $W_{e} \in \mathbb{R}^{h \times 3h}$, $s_j, e_j \in \mathbb{R}^{h}$ and $\eta^j \in \mathbb{R}^{L}$.
	
	The model update historical and current information iteratively for further reasoning. 
	We use the start and end evidence vector $s_{j-1}$,\, $e_{j-1}$ of the current block as the initial states of the subsequent block. Circularly, we obtain the updated representation $\tilde{U}$ (Fig. \ref{fig:gate} - \textcircled{\small{5}}).
	
	\begin{equation}\label{equ:concat}
	\tilde{u}_l = {\rm ReLU}(W_{a} [s_{j+1} \otimes \mathbf{e}_L; e_{j+1} \otimes \mathbf{e}_L; u_l])
	\end{equation}
	
	In order to make different decisions in different situations, we introduce the gate mechanism to determine the stimulus intensity of reverse thinking in memory. We take the decoded tokens $y_t$ into account, the attention $\alpha_t$ comes from the multi-head attention \cite{DBLP:conf/nips/VaswaniSPUJGKP17}, the ${v}^{new}_{m}$ is defined as:
	\begin{equation}\label{equ:vm}
	{v}^{new}_m = \sum^T_{t}\alpha_t y_t
	\end{equation}
	Where $T$ is the length of the outputs of the decoder. Then, the context vector $c_t$ which is a part of the input to the gate modeled by max-pooling over ${v}^{new}_m$ $M$ multiple times, as: 
	\begin{equation}\label{equ:ct}
	c_t = {\rm max\verb|-|pooling}(\left\{v^{new}_m\right\}_{m=1}^M) \in \mathbb{R}^{h}
	\end{equation}
	
	Finally, the Gate-Reverse Thinker yield the output ${u}^{new}_l$ (Fig. \ref{fig:gate} - \textcircled{\small{8}}) with the $\tilde{u}_t$ (Eqn. (\ref{equ:concat})) and the gate vector $g_l$ that can be formulated by the context vector $c_t$ and $u_t$ (Eqn. (\ref{eq:bert})), as:
	\begin{align}\label{equ:gt}
	g_l = \sigma(W_g u_l& + W_g' c_t) \in \mathbb{R}^{1}	\\
	\label{unew}
	{u}^{new}_l &= g_l \circ \tilde{u}_l
	\end{align}
	Where $W_g \in \mathbb{R}^{h}$, $W_g' \in \mathbb{R}^{h}$ and $\sigma$ represents a sigmoid function, $\circ$ indicates element-wise multiplication.
	
	{\bf Fusion Layer}:
	To combine the reverse thinking and inertial thinking, we adopt the fusion kernel used in \newcite{DBLP:conf/acl/WangWY18} for better semantic understanding:
	\begin{equation}\label{equ:fusion}
	r_l = {\rm {Fuse}}(\alpha{u^*}^{new}_l, \beta{u}^{new}_l)
	\end{equation}
	Where ${u^*}^{new}_l$ is the output of No-Gate-Inertial Thinker described in Sec. (\ref{retain}), but here, it is initialized with a random vector in the first training step. $\alpha$ and $\beta$ are hype-parameters.
	
	{\bf Soft-Decoder}: Following the paper \cite{DBLP:conf/nips/VaswaniSPUJGKP17}, we use a stack of Transformer decoder blocks on top of the embeddings provided by the word embedding layer and self attention. The second and third sub-layers perform the multi-head attention over $V$ (Eqn. \ref{v}) and $r_l$ (Eqn. \ref{equ:fusion}). Also, the pointer-softmax mechanism \cite{DBLP:conf/acl/GulcehreANZB16} that learns to switch between copying words from the document and generating words from a prescribed vocabulary is used. We do not describe the details here due to space limitation, instead directly give the probability distribution of the $t$-th token, as:
	
	\begin{equation}\label{equ:decoder}
	p(y_t | y_1, \ldots ,y_{t-1}; \theta) = {\rm {Soft\verb|-|Decoder}}({V}; r_l)
	\end{equation}
	
	\subsection{Retraining with Inertial Thinking}
	\label{retain}
	
	To satisfy the situation described in Sec. \ref{clst}, we retrain our model on the condition of the reverse thinking in a positive way.
	
	Given the passage $X^P$ and question $X^Q$, the Forward Encoder, having different parameters with the Backward Encoder, encodes the contextual information $u'_l$ and $\tilde{V'}$ that is similar to Eqn. (\ref{eq:bert}; \ref{V}; \ref{v}). And the No-Gate-Inertial Thinker builds the obverse logical relationship in a multi-step reasoning (Fig. \ref{fig:gate} \fbox{box}) fashion without a gate, like Eqn. (\ref{equ:concat}): 
	\begin{equation}\label{for:thinker}
	\tilde{u}'_l = {\rm No\verb|-|Gate\verb|-|Inertial~Thinker} (\tilde{V'}; u'_l)
	\end{equation}
	
	Then, Gate-Reverse Thinker outputs the reverse thinking information based on $\theta$, like Eqn. (\ref{unew}):
	\begin{equation}\label{for:thinker-back}
	{u'}^{new}_l = {\rm Gate\verb|-|Reverse~Thinker} (\tilde{V'}; {\tilde{u}}'_l; \alpha'_t; y'_t|~\theta)
	\end{equation}
	
	After calculating the inertial $\tilde{u}'_l$ and reverse ${u'}^{new}_l$ vectors, similarly, we obtain the bi-directional knowledge $r'_l$ like Eqn. (\ref{equ:fusion}) and use it to decode the probability distribution of the $t$-th answer, as:
	\begin{equation}\label{for:fusion}
	\begin{gathered}
	r'_l = {\rm {Fuse}'}(\alpha\tilde{u}'_l, \beta{u'}^{new}_l)	\\
	p(y'_t | y'_1, \ldots ,y'_{t-1}; \theta') = {\rm {Soft\verb|-|Decoder}}'({V}'; r'_l)
	\end{gathered}
	\end{equation}
	where $\alpha$ and $\beta$ are manual specified which determine the proportion of the bi-directional thinking.
	
	
	
	
	\section{Experimental Setup}
	\subsection{Dataset \& Evaluation Metrics}
	To demonstrate the effectiveness of our work, we choose DuReader benchmark dataset which is designed from real-world search engines (BaiDu). In terms of the data size, it contains 300k questions and the data has been split into a training set (290k pairs) and a development set (10k pairs). The test split of DuReader is hidden from the public. Therefore, we take 5k question-answer pairs randomly from the development data as validation set and use the rest development data to report test results. As for the evaluation metrics, the answers are human-generated and not necessarily sub-spans of the passages so that the metrics in DuReader are {ROUGE-L} (R-L) \cite{lin2004rouge} and {BLEU-4} (B-4) \cite{DBLP:conf/acl/PapineniRWZ02}.
	
	\subsection{Implementation Details}
	The BERT-style baselines have the same hyperparameters given on the paper \cite{DBLP:journals/corr/abs-1907-11692}. We use Adam optimizer \cite{DBLP:journals/corr/KingmaB14} for training, with a start learning rate of 3e-5 and a mini-batch size of 32. The epoch of DuReader dataset is 12. To coordinate the bi-directional knowledge, we set $\lambda 1$ and $\lambda 2$. 
	To improve both the efficiency of training and testing, following \newcite{DBLP:conf/acl/HeLLLZXLWWSLWW18}, we select one most related paragraph from labeled documents as input. $\alpha$ and $\beta$ are set as 0.8 and 0.2, respectively.
	
	\section{Results}
	In this section, we evaluate our model on the benchmark dataset: DuReader dataset \cite{DBLP:conf/acl/HeLLLZXLWWSLWW18}. Pre-trained language model RoBERTa (RB) and some state-of-art models are used as baselines to test the performance of the proposed BCTN. We use the large pre-trained language model (-large) in the main experiments for getting a strong result. However, to increase training speed, the base model (-base) are utilized in our ablation study and other experiments. The experimental results and analyses validate the effectiveness of our model.
	
	\newcommand{\unpublished}{$^*$}
	
	\begin{table}[!t]
		\centering
		\resizebox{0.9\columnwidth}{!}{
			\begin{tabular}{lcccc}
				\toprule
				Model & Dev ROUGE-L & Dev BLEU-4 & Test ROUGE-L & Test BLEU-4\\
				\midrule
				\textit{Extracted model} \\
				Match-LSTM \cite{DBLP:conf/iclr/Wang017a}                   & 32.36 \unpublished & {39.67} \unpublished &{31.80} & {39.00} \\
				BiDAF \cite{DBLP:conf/iclr/SeoKFH17}    & 32.75 \unpublished  & 40.38 \unpublished & 31.90 & 39.20 \\
				V-Net \cite{DBLP:conf/acl/WuWLHWLLL18}                     & $-$  & $-$ & 44.18 & 40.97 \\
				R-Net \cite{DBLP:conf/acl/WangYWCZ17} & $-$ & $-$ & 47.71 & 44.88 \\
				Deep Cascade \cite{DBLP:conf/aaai/YanXWBZZSWWC19} & $-$ & $-$ & 50.71 &  \textbf{49.39} \\
				\midrule
				\textit{Generative model} \\
				RB-Base     & 55.32  & 39.78 & 54.18 & 38.85 \\
				BCTN-Base  & 58.52 & 43.86 & 58.04 & 43.19 \\
				\midrule
				RB-Large     & 57.44  & 42.08 & 56.86 & 41.87 \\
				BCTN-Large      & \textbf{59.90}  & \textbf{45.26} & \textbf{59.12} & 44.53 \\
				\midrule
				Human	\cite{DBLP:conf/acl/HeLLLZXLWWSLWW18}      & $-$  & $-$ & 57.4 & 56.1 \\
				\bottomrule
		\end{tabular}}
		\caption{
			ROUGE-L and BLEU-4 scores (\%) on DuReader test set for single models.
			\unpublished~ indicates that the performance is reimplemented by
			ourselves.
			$-$ indicates that the development scores were not publicly available at the time of writing.
		}
		\label{table:perf-leaderboard}
		
	\end{table}
	
	\subsection{Main Experiment}
	In DuReader dataset, baselines can be classified as three types: state-of-the-art models, RoBERTa-base (RB-base) model and RoBERTa-large (RB-large) model. RB-base and RB-large indicate that we directly use the pre-trained language model as the encoder without MTL System. In order to reduce the complexity of the model, previous methods \cite{DBLP:conf/iclr/SeoKFH17,DBLP:conf/acl/WuWLHWLLL18,DBLP:conf/aaai/YanXWBZZSWWC19} turn the DuReader into extractive tasks. Therefore, we divide the models into extractive models and generative models. As shown in Table \ref{table:perf-leaderboard}, the main results of our single model on DuReader outperforms the BERT-style baselines, 3.86\% gain on ROUGE-L and 4.34\% gain on BLEU4 on the RoBERTa-base model, as well as 2.26\% gain on ROUGE-L and 2.66\% gain on BLEU-4 on the RoBERTa-large model. Although our model has decreased a little on BLEU-4 compared to the extractive models, it outperforms them on ROUGE-L about 8.4\%. And extractive models usually have better performances.
	
	\subsection{Ablations Study \& Effect of Different Parameters}
	We conducted an ablation study on our model to discuss the impacts of the augmented components which can be removed in our framework. Table \ref{tab:ablation} shows the effectiveness of different parts in our proposed BCTN. Note that by removing all different elements, configuration 3 reduces to the RB-base model.
	
	
	\begin{figure*}[t]
		\begin{minipage}{0.48\textwidth}
			\centering\small
			\setlength{\tabcolsep}{2pt}
			\scalebox{1.0}{
				\begin{tabular}{lcc}
					\hline
					{\textbf{Method}} & {\textbf{ROUGE-L(\%)}} & {\textbf{BLEU-4(\%)}} \\ \hline
					\textbf{Complete Model} & \textbf{58.04} & \textbf{43.19}  \\ \hline
					1) w/o Gate Mechanism & 56.99 & 41.89  \\
					2) w/o Gate-Reverse Thinker & 57.30 & 39.05  \\ 
					3) w/o MTL System & 54.18 & 38.85  \\ \hline
				\end{tabular}}
			\captionof{table}{\label{tab:ablation} Ablation study results on model components. Performance of different configurations on the DuReader dataset.}
			\label{tab:learning_procedures}
		\end{minipage}
		\hfill
		\begin{minipage}{0.48\textwidth}
			\vspace{0pt}
			\centering
			\small
			\setlength{\tabcolsep}{2pt}
			\scalebox{1.0}{
				\begin{tabular}{lcc}
					\hline
					~~\textbf{Method} & {\textbf{ROUGE-L(\%)}} & {\textbf{BLEU-4(\%)}}   	\\
					\hline
					~~RB-base 	& 54.18  & 38.85 	\\ 
					\hline
					~~BCTN ($\alpha$=1, $\beta$=0)	& 57.02  & 40.55 \\
					~~BCTN ($\alpha$=0.8, $\beta$=0.2) & \textbf{58.04} & \textbf{43.19}\\ 
					~~BCTN ($\alpha$=0.6, $\beta$=0.4) & 54.64 & 33.07\\
					~~BCTN ($\alpha$=0.4, $\beta$=0.6) & 54.08 & 35.88\\ 
					~~BCTN ($\alpha$=0.2, $\beta$=0.8) & 46.85 & 18.64 \\
					~~BCTN ($\alpha$=0, $\beta$=1) & 20.68 & 2.12 \\
					\hline
				\end{tabular}}
			\captionof{table}{Results of BCTN variants and baseline.} 
			\label{exp:parameters}   
		\end{minipage} %
	\end{figure*}
	
	According to Table \ref{tab:ablation}, the R-L and B-4 of origin system is 54.18\% and 38.85\%. And we can conclude that: 1) the Gate Mechanism which determines the stimulus intensity of reverse thinking in memory is effective to the proposed model. 
	2) To validate the importance of reverse thinking, we remove the Backward Encoder, which results in a performance loss, about 4.14\% on BLEU-4. 3) The medial temporal lobe system makes a contribution to the overall performance, which confirms our hypothesis that combining the bi-directional knowledge to answer questions is necessary. Either component is important for our proposed model.
	
	We manually set the different parameters $\alpha$ and $\beta$ (Eqn. \ref{equ:fusion}) to explore how the bi-directional knowledge affects the performance of the BCTN. From the Table \ref{exp:parameters}, we can observe that the performance of the model reaches 57.02\% on ROUGE-L when only using the inertial thinking, however, the model achieves a peak after adding reverse thinking. In the case where the model uses only reverse thinking and ignores inertial thinking, the model's effectiveness drops significantly. This is consistent with human behavior in psychology that reverse thinking can assist inertial thinking to generate more accurate answers and it is not enough to use reverse thinking or inertial thinking alone.
	 
	\subsection{Qualitative Examples}
	Qualitatively, we have observed interesting examples in DuReader before and after adding the bi-directional thinkers. As shown in Table 3, in case one, the proposed model output a generative question \textit{How to pass the master level of "End of Nightmare"} which has the same semantic with the gold question. And our proposed BCTN gets the correct answer with a more detailed explanation that marks blue. However, the RB-base output a wrong answer with the ambiguous response, especially the sentence \textit{they must reach the blood}. In case two, the same conclusion can be reached too. The answer of RB-base describes the nutrients of the loquat instead of the real question. But the BCTN not only gives the response correctly, but also explains why can pregnant women eat loquat. The answers become more explicable and proper with the help of our model, illustrating that 
	our ideas can indeed assist question answering.
	
	\begin{table*}[t!]
		\centering
		\label{tab:cherrypick}
		\begin{tabular}{rp{0.83\textwidth}}
			\toprule
			Document & \textit{{\color{green}Players} who want to {\color{green}clear the master level of ``End of Nightmare" need to kill the hero Bian Que.} If you want to kill Bian Que, you must {\color{red}arrive at} Bian Que {\color{red}before the Iron Face Man}, because the Iron Face Man is invincible in this level. In the level, players should have as few contacts with monsters as possible ...} \\
			$\mathrm{Q_{gold}}$ & \textit{How to get through "End of Nightmare" (Master Difficulty) ?} \\
			$\mathrm{Q_{gen}}$ & \textit{How to pass the master level of "End of Nightmare" ?} \\
			Answer & {\tt RB-base}: \textit{Players want to clear the level, {\color{red}they must reach the {blood}} before the Iron Faced Man.}	\\
			& {\tt BCTN:} \textit{{\color{green}To clear the master level at the end of the nightmare, you need to kill the hero Bian Que.} {\color{blue}If you want to kill Bian Que, you must arrive at Bian Que before the Iron Face Man, because the Iron Face Man is invincible in this level.}} \\
			\midrule
			Document & \textit{Loquat is a kind of southern fruit ...... Loquat has a unique flavor and is rich in nutrients. {\color{green}Pregnant women can eat it.} It contains protein, {\color{red}fat}, fruit acid, {\color{red}fructose and calcium, phosphorus, iron, sodium, potassium and other minerals}. {\color{green}Pregnant women eating loquat can increase appetite, relieve heat and relieve thirst.} Loquat can stimulate the secretion of digestive glands, and has a good effect on increasing appetite, helping digestion and absorption, and quenching thirst and relieving heat ... } \\
			$\mathrm{Q_{gold}}$ & \textit{Loquat, can pregnant women eat it ?} \\
			$\mathrm{Q_{gen}}$ & \textit{Can pregnant women eat loquat ?} \\
			Answer & {\tt RB-base}: \textit{{\color{red}Fructose, fat, calcium, phosphorus, iron, sodium, potassium.}}\\
			& {\tt BCTN:} \textit{{\color{green}Pregnant women can eat loquat which increases appetite, relieves heat and relieves thirst.} {\color{blue}Loquat can stimulate digestive gland secretion, and have a good effect on increasing appetite, helping digestion and absorption.}} \\
			\bottomrule
		\end{tabular}
		\caption{Two examples of the results in DuReader. Gold answers correspond to text in green. The red text indicates the wrong answers or not exactly. In both the cases, the answers produced by the BCTN are highly related with the question. And the generated question is similar with gold question.}
	\end{table*}
	
	\section{Conclusion}
	In this paper, we present the Bi-directional Cognitive Thinking Network (BCTN) which corresponding to the Bi-directional Cognitive Knowledge Framework (BCKF) from the perspective of psychology. The BCTN answers the question with bi-directional knowledge by simulating the inertial thinking and reverse thinking. And we decouple these two parts of knowledge rather than couple them with the same module.
	To determine the stimulus intensity of reverse thinking in memory, we consider the decoded tokens to calculate the score based on the gate mechanism. We show that the proposed BCTN is very effective, it has competitiveness with the previous methods in literature on DuReader with a single model. Our future work will consider to use different datasets and design various models to simulate the behavior of our brain to capture human-level language understanding and intelligence. Finally, we believe that our framework can generalize to other generative tasks,
	such as summarization and image caption.
	
	\section*{Acknowledgements}
	We thank all anonymous reviewers for their constructive comments. This work is supported by the National Natural Science Foundation of China (No.62006222). 
	
	\bibliographystyle{coling}
	\bibliography{coling2020}

\end{document}